# Integration and Implementation Strategies for AI Algorithm Deployment with Smart Routing Rules and Workflow Management

Barbaros Selnur Erdal[1], Vikash Gupta[1], Mutlu Demirer[1], Kim H. Fair[1], Richard D. White[1], Jeff Blair[2], Barbara Deichert[2], Laurie Lafleur[2], Ming Melvin Qin[3], David Bericat[3], Brad Genereaux[3]

[1] Mayo Clinic Florida | [2] Laurel Bridge | [3] NVIDIA

List of Abbreviations in order of appearance:

PACS: Picture Archiving and Communication System
AI: Artificial Intelligence
DICOM: Digital Imaging and Communication in Medicine
HL7: Health Level 7
FHIR: Fast Healthcare Interoperability Resources
IHE: Integrating the Healthcare Enterprise
MONAI: Medical Open Network in Artificial Intelligence
HIS: Hospital Information System
RIS: Radiology Information System
VNA: Vendor Neutral Archive
AIR: Artificial Intelligence Results
DICOM SR: DICOM Structured Report
AIW-I: Artificial Intelligent Workflow in Imaging
CDSS: Clinical Decision Support System
MAP: MONAI Application Package
HIPAA: Health Insurance Portability and Accountability Act
GDPR: General Data Protection Regulation
EMR: Electronic Medical Records
AE Title: Application Entity Title

# Abstract

This paper reviews the challenges hindering the widespread adoption of artificial intelligence (AI) solutions in the healthcare industry, focusing on computer vision applications for medical imaging, and how interoperability and enterprise-grade scalability can be used to address these challenges. The complex nature of healthcare workflows, intricacies in managing large and secure medical imaging data, and the absence of standardized frameworks for AI development pose significant barriers and require a new paradigm to address them.

The role of interoperability is examined in this paper as a crucial factor in connecting disparate applications within healthcare workflows. Standards such as DICOM™, Health Level 7 HL7®, and Integrating the Healthcare Enterprise (IHE) are highlighted as foundational for common imaging workflows. A specific focus is placed on the role of DICOM gateways, with Smart Routing Rules and Workflow Management leading transformational efforts in this area.

To drive enterprise scalability, new tools are needed. Project MONAI, established in 2019, is introduced as an initiative aiming to redefine the development of medical AI applications. The MONAI Deploy App SDK, a component of Project MONAI, is identified as a key tool in simplifying the packaging and deployment process, enabling repeatable, scalable, and standardized deployment patterns for AI applications.

The abstract underscores the potential impact of successful AI adoption in healthcare, offering physicians both life-saving and time-saving insights and driving efficiencies in radiology department workflows. The collaborative efforts between academia and industry, are emphasized as essential for advancing the adoption of healthcare AI solutions.

# Introduction

Computer vision applications adapted for the specifics in the healthcare industry are being conceptualized, developed, validated, and deployed into hospital ecosystems, serving both insights to physicians that can be lifesaving and timesaving, as well as driving efficiencies in radiology department workflows.

However, unlike other industries that have been transformed by Artificial Intelligence (AI), mass adoption of healthcare AI solutions remains an elusive problem. The reasons are as follows:
- Healthcare workflows require significant orchestration between the core applications that are produced by distinct manufacturers.
- Medical imaging data can be challenging to manage due to its large size, complex structure, and security requirements.
- The lack of standardized frameworks for developing medical AI applications makes the software delivery lifecycle brittle, complex, and expensive.

Interoperability and enterprise-grade scalability are the keys to addressing these issues. Interoperability acts as a glue between these disparate applications and is effective for the common imaging workflows (order, protocol, acquisition, review, report, and distribution). Standards exist that support these workflows, with DICOM™ [1] and Health Level 7 HL7® [2] at the forefront and Integrating the Healthcare Enterprise (IHE) [3] providing defined workflows. A DICOM gateway [4] is one of the most ubiquitous components in this workflow and is enabling the transformation now.

To drive enterprise scalability, a new paradigm in developing AI applications is needed. Project MONAI [5], the Medical Open Network for Artificial Intelligence, was established in 2019 to establish this paradigm, amongst both academia and industry, including NVIDIA [6]. MONAI Deploy App SDK [7], one of the components of MONAI, elevates the packaging process of taking AI models and instantiating AI applications. As a framework for creating medical AI applications, MONAI Deploy App SDK enables repeatable, scalable, and standardized deployment patterns for applications, simplifying the work by IT and DevOps teams to support AI models as they transition from research into production usage [8, 9].

## A World Born of Interoperability

In the latter part of the prior century, producing and consuming devices of digital medical imaging data found it difficult to communicate with one another. A standard for medical imaging communication was not only desired but necessary for the adoption of digital imaging in hospitals. Born out of this necessity in the 1980s was the DICOM standard, and since then, it has been universally adopted as the standard of choice for medical imaging. While DICOM has solved many problems, interoperability scenarios still exist such as compression format mismatches, implementation mismatches between client and servers, and missing required information. In addition, as the DICOM protocol is built for the medical imaging domain, concepts and paradigms may be daunting to the AI developer. However, leveraging the DICOM standard for each unique AI algorithm correlates to faster time to market, faster adoption, and a more flexible and scalable overall solution.

For healthcare practices, DICOM adoption was pivotal. As an analogy, consider the paradigm shift that happened when the contents of large filing cabinets were digitized and stored electronically. New file standards emerged, such as Microsoft Word, Excel, PowerPoint, and Adobe PDF, along with data sharing protocols between different systems and different networks. As connectivity increased, new standards for file sharing and storage emerged. One of the major impacts of digitization and data sharing is a better understanding of localized statistical data analysis that leads to insights for making data-driven policy changes.

A parallel movement is happening in the field of medical images and electronic health records (EHR). As noted above, DICOM emerged as the imaging and sharing standard since its introduction in the 1980s. The DICOM standard along with other healthcare standards such as HL7's v2 and FHIR® have made it possible to connect different clinical infrastructure starting from radiologist's workstation to PACS. There has been localized data driven research, but healthcare is far from reaching the "Big Data" scenario that has been seen in other industries.

Because of the use of these imaging standards, AI models can now be trained that can impact patient care at an individual level. Only very recently could AI algorithms for text and images be used at an institutional level for making large policy changes. With the introduction of tools like ChatGPT and Generative AI, individuals are using AI for solving personalized problems. A similar shift can be expected in medical imaging AI, where AI models can be used for reading individual cases. This is being achieved in some capacity across hospitals, but not yet at large scale.

Providing a large-scale personalized AI solution is a major IT challenge for hospitals and PACS administrators. Imaging informaticists struggle with managing the breadth of AI models and applications, getting data to and from these applications, visualizing the results from these applications, and managing the sheer scale necessary for healthcare operations. This paper systematically reviews one such implementation at the Center for Augmented Intelligence in Imaging (CAII) at Mayo Clinic, Florida. It looks at the MONAI Model Zoo [10] for AI application discovery, DICOM Routing capabilities for managing both the data to and from the applications as well as supporting enterprise scale, and Mayo Clinic's CAII Viewer for visualizing the AI results.

## The Paradigm for Next Generation AI Integration

Mayo Clinic in Florida combined the usage of multiple tools including MONAI, DICOM Router functionalities, and in-house developed CAII Viewer to achieve deep AI integration at scale.

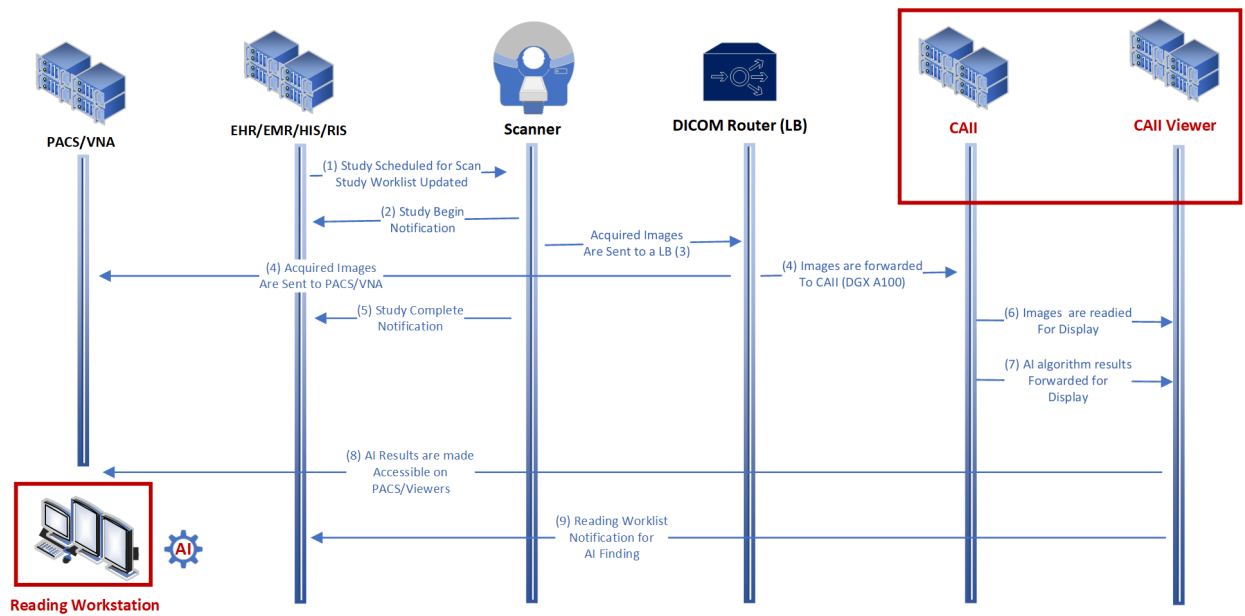

*Figure 1: Basic Radiology Workflow Steps.*

Figure 1 shows a basic radiology workflow that shows an order being generated, image-data being acquired during patient scanning, produced images being evaluated by a radiologist, and a report being generated by the image interpreter and sent back to the clinician for review. Table 1 describes these steps in detail.

| Step | Description |
|---|---|
| 1 | When the Clinical Team orders an imaging study, an exam is scheduled. This triggers an update on Study Worklist (aka. Modality Worklist). |
| 2 | When the scheduled exam is worked on by the Imaging Team, a "begin exam" message is sent to HIS/RIS. |
| 3 | The acquired images are sent to DICOM Router. |
| 4 | DICOM Router sends the acquired images to PACS / VNA, as well as a copy to the CAII Servers. |
| 5 | When the full image acquisition is completed, a "study complete" message is sent to the HIS/RIS. |
| 6 | CAII makes the images ready to be viewed (currently on NVIDIA DGX A100[11]). |
| 7 | CAII AI algorithms (also running on a NVIDIA DGX A100) run inference on the forwarded images and make them available on the CAII Viewer (Figure 2). These results follow AIR [12] standards (e.g., TID1500 for DICOM SR generation, etc.). |
| 8 | CAII Viewer makes imaging studies (identified by an accession number) and any available AI results for them viewable/accessible through a URL hyperlink. |
| 9 | It is also possible for an AI Algorithm to send a priority message to a worklist to drive attention to a specific critical finding, etc. |

*Table 1: Steps in Workflow.*

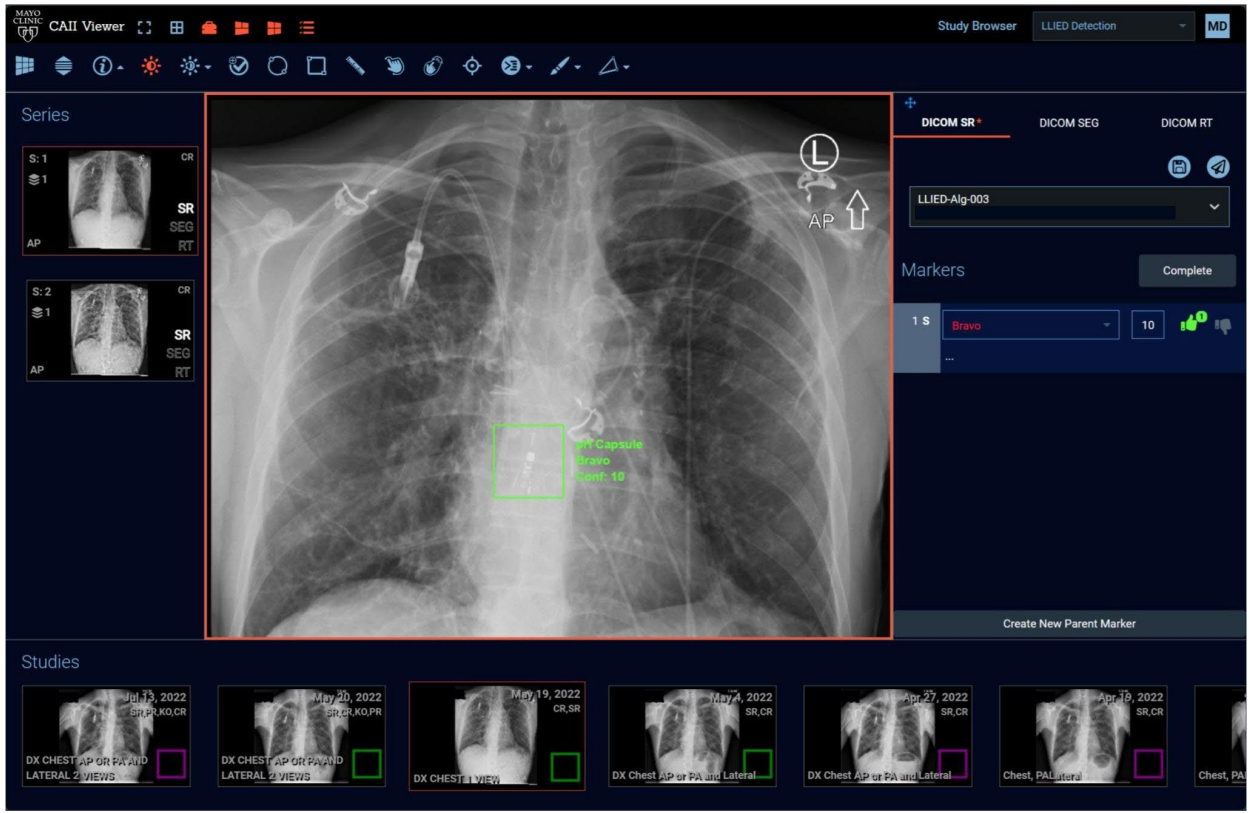

*Figure 2: MRI-unsafe device (Bravo esophageal reflux pH capsule) correctly detected and identified (with 10/10 certainty) on a chest x-ray by model inference (shown as a solid bounding box) displayed and adjudicated by a radiologist on the viewer. The radiologist can view previous exam results as well. [13]*

Table 2 highlights the common workflow themes in a typical radiology workflow, and what is discussed in this paper. Steps covered in this paper are indicated by an asterisk. The IHE AI Interoperability in Imaging white paper [14] provides examples of AI applications that could be used for these themes, and IHE's AI Workflow for Imaging (AIW-I) profile [15] provides specific interoperability specifications to drive actor and transaction communication.

| Step | Common Workflow Theme | IHE AI White Paper Described Boundaries |
|---|---|---|
| **1. ORDER** | When a clinician orders an imaging examination in the HIS/RIS, they may be guided by a CDSS to ensure its appropriateness. Depending on the clinical setting, the order may contain a clinical-status priority code (e.g., "stat"). | Make recommendations as to the types of procedures that should be ordered, based on the patient's condition and record. |
| **2. PROTOCOL** | Once the patient examination is scheduled for a date and location, an entry is created on the "study worklist" of the scanner (or another imaging device). In some instances, an entry is also created on a "protocoling worklist", where a radiologist determines the specific imaging techniques to be used (e.g., scanning details, contrast-agent type / amount / administration route) during the diagnostic imaging study or image-directed procedure. | Make recommendations on the type of protocol to be used on the scanner. |
| **3. POST-PROCESS** | Once the examination is completed, images are reconstructed into a human-interpretable format and sent to a DICOM router to be forwarded to the appropriate destinations, including a PACS and/or VNA for management or storage. Once the organized images (original and/or post-processed) are ready to be evaluated by the radiologist, the examination description appears on the radiologist's "reading worklist". | Post-process the image, identify quality assurance (QA) issues prior to the patient leaving the department, and prepare classifications and segmentations in advance of the radiologist's evaluation. |
| **4. READ *** | Radiologists assess the examination images on their diagnostic viewer and dictate their interpretation (typically into a voice recognition system). | Include insights alongside the images in the radiologist's display. |
| **5. REPORT *** | The dictated report is sent to the HIS/RIS. If actionable critical and/or non-critical findings are identified, radiologists may invoke additional workflows to alert the ordering clinician and issue the final examination report. | Include emergent insights for consideration of the ordering physician. |
| **6. DISTRIBUTE *** | Final examination reports become available in the HIS/EHR, along with the images in the PACS or clinical viewers. | Pre-populate the radiologist's report with draft insights to be considered by the radiologist. |

*Table 2: Steps in a typical radiology workflow. Steps marked by an asterisk (*) are the areas of focus for this paper.*

## Building an AI Integration Strategy is Essential

Workflow integration into the medical imaging ecosystem is key to having a transformative impact. Smart Routing Rules and Workflow Management streamlines this integration by providing tools for medical image routing and workflow management. MONAI helps by providing tools for integrating medical AI applications into existing workflows, such as communicating with medical imaging systems and electronic health records (EHRs) systems.

It takes an interdisciplinary team to create an AI-enabled medical ecosystem. Radiologists, technologists, PACS administrators, IT personnel, and data scientists must come together to guide, build, and sustain AI-powered workflows. Smart Routing Rules and Workflow Management helps to facilitate collaboration by providing tools for sharing medical images and other data securely. MONAI helps by providing a platform for collaboration and sharing of medical AI applications and resources, including pre-trained models, algorithms, and datasets.

Medical imaging applications must comply with regulatory requirements, including HIPAA and GDPR privacy regulations, and these have implications on the data connectivity layer. Smart Routing Rules and Workflow Management can help ensure regulatory compliance by providing tools for secure image transfer and DICOM data management. MONAI can also help by providing tools for ensuring regulatory compliance during algorithm development and model deployment.

By combining the strengths of DICOM Routing and MONAI, medical AI developers can develop and deploy enterprise-scale medical imaging and AI applications that are seamlessly integrated with existing imaging workflows and systems.

## Smart Routing Rules, Workflow Management and MONAI: Better Together

Building MONAI AI applications atop Smart Routing Rules and Workflow Management as the integration "glue" comes with many benefits for the healthcare organization. This includes the ability for parallel processing, managing deviations from the DICOM standard, and enhanced routing flexibility.

DICOM transport pipelines can be tapped to allow for parallel processing. A traditional workflow might have a modality send an imaging study to a post-processing system, and then onward to PACS, which may in turn send that to several AI algorithms in serial fashion. Doing so introduces a lot of latency when post-processing and AI algorithm execution could occur simultaneously on different systems. Utilizing an engine like Laurel Bridge allows for parallel routing so that this analysis can occur simultaneously by different systems, reducing latency and saving time.

Some image processing libraries have some brittleness when it comes to handling non-conformant DICOM. For example, DICOM private tags could be problematic for some processing engines not capable of gracefully ignoring these fields. A modality claiming that the pixel data is uncompressed but sends compressed data may cause libraries to fail. Smart Routing Rules and Workflow Management provides the opportunity to adjust content in flight with features like tag morphing.

Coordinating the ever-growing set of AI algorithms healthcare systems will be faced with – Smart Routing Rules and Workflow Management provides a one-stop to configure endpoints in a solution that focuses on that interoperable layer. This is important for managing different environments (development, testing, pre-production, and production) – and being able to roll forward and roll backward updates and change IP address/port/AE titles with ease is important.

Adding additional routing logic via a user interface in a DICOM router (LB) may be easier than in a MONAI pipeline. For example, setting up a workflow that will send all imaging studies from "modality 1" but blocking all imaging studies from "modality 2", is better suited at the routing layer, rather than sending everything to the AI application only to be tossed away there. Another example is that of thin and thick slice series; if an AI algorithm only needs thick slice series, it may be wasteful to send the larger thin-slice series to the AI algorithm, clogging both the networks and transitional storage. DICOM routers can suppress those to reduce the burden of network traffic.

## How to Get Started

To deploy a model in hospital infrastructure, a model can be trained on a representative dataset. To start, sample models from the MONAI Model Zoo can be retrieved. There are multiple methods to deploy a MONAI MAP model, as described in Table 3.

| Method | Description |
| --- | --- |
| 1 | Build a MONAI Application Package (MAP) using an in-house developed model. |
| 2 | Build a MONAI model bundle and wrap it in a MAP. |
| 3 | Download an existing MAP and deploy it. |
| 4 | Deploy the MAP and building routing rules in DICOM Router. |

*Table 3: Methods to deploy a MONAI AI application.*

This paper describes method 4, leveraging a DICOM Router to deeply integrate MONAI MAPs into clinical workflows. This approach can be summarized in pseudo-code described in Figure 3.

```
LB AI-Results-Handling (Image Studies)

    Check each exam for matching exam descriptions

    If (matching exam found) {

      /*LB and AI algorithm inside the institutional firewall*/

        LB to DICOM-Send to AI-Receiver

        /*Results in the form of DICOM SC and DICOM SR*/

        AI-Receiver to return results to LB

        LB to Forward the DICOM SC, SEG, RT or DICOM SR to Viewers

        /*Using Custom C# based DICOM SR parsing to produce an XML template to

        provide mapping during HL7 conversion by LB*/

        LB to Produce HL7 from DICOM SR

        LB to Send HL7 message to Interface Engine for HIS for Prioritization
```

*Figure 3: Pseudo-code for DICOM Router (Laurel Bridge: LB) integration with MONAI MAP*

If the given finding is expected to trigger a prioritization message, LB enables custom C# codes to be executed to produce HL7 messages (driven from DICOM SR TID1500 messages) to be forwarded to various worklists. Described in figure 4 is a sample HL7 v2 message for EMR integration.

```
MSH|^~\&|AI Product Name{MONAI_TEST} | HIS{HIS_TEST}| |Timestamp| |ORM^O01|GUID|T|2.5.1

PID|||patient ID^^^MC^MC||Patient last name^Patient first name||Patient DoB

ORC|XO||||||||Datetime of transaction

OBR|1||Accession number|Study code^Study description^IMAGEID^^Short study

description|||Study Date|||||||||||||Study Date

OBX|1|ST|AI_PRIORITY_type of evaluation{AI_PRIORITY_MONAI}||priority level{HIGH}

OBX|2|ST|AI_DETECTION_type of evaluation{AI_DETECTION_MONAI}||detection{POS}
```

*Figure 4: Sample HL7 v2 message for EMR integration.*

# Building a MONAI Application Package

A MONAI Application Package (MAP) is common to almost all deployment scenarios, and as such, it is important to understand how it is constructed. A MAP is a chain of MONAI Operators connected in a Directed Acyclic Graph (DAG) manner. Operators are designed to perform a single function, like applying a Gaussian filter, thresholding an image, or making predictions on an image using a deep learning model. See Figure 5. It is also important to note that a MAP can be any collection of operators and does not need to be only based on deep learning concepts. A MAP can be triggered either with Python or by using a "docker exec" command. In addition, a MAP also provides a command line interface (CLI) for executing the application.

MONAI Deploy App SDK also comes with built-in operators for parsing incoming DICOM data. Operators like "StudyLoaderOperator", "SeriesSelectorOperator", and "SeriesToVolumeOperator" are chained together in the above-mentioned order for converting the DICOM study to a "numpy" array which can then be used for inference. These are generic DICOM readers. However, if these operators are not able to read the given DICOM studies, custom operators can be developed and added to the application. The outputs with the AI results need to be written in a format consumable to the clinical viewers, like DICOM Segmentation or DICOM Structured Reports formats.

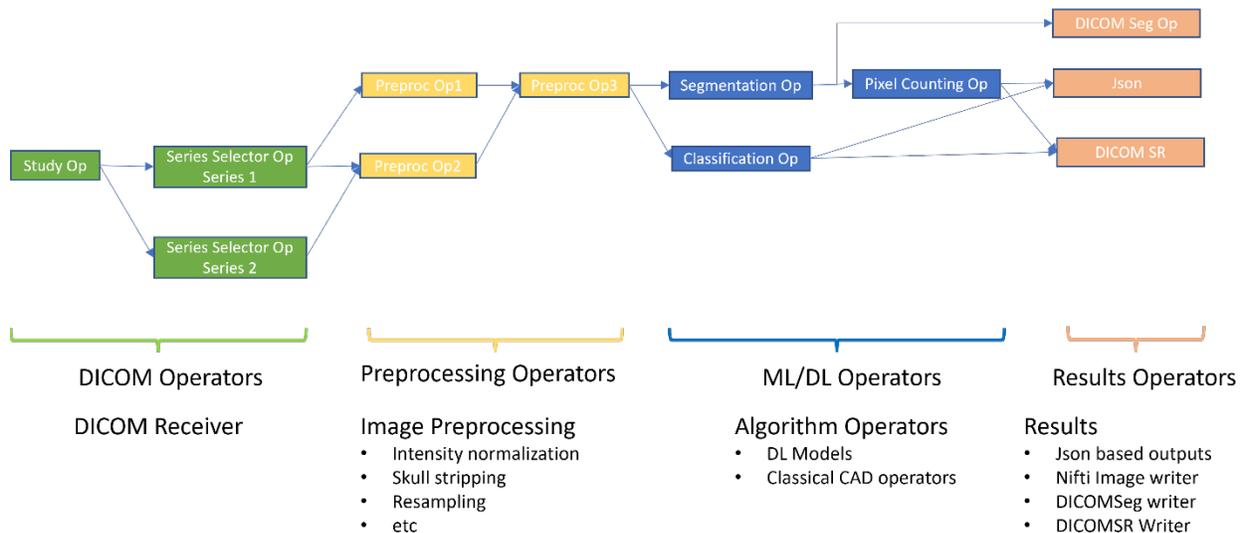

Figure 5: The MONAI Deploy image processing pipeline.

# Configurability of Smart Routing Rules

Utilization of DICOM routing rules, divides configurability into 3 logical groups: the sources, the rules processing, and the destinations. The source logical group allows for configuration of the type of source, i.e., file folder input, DICOM, or DICOMweb, etc., in addition to various connectivity specifics. The rules engine allows for fine grained control of how to handle the imaging data that it receives. This includes designation of a destination or set of destinations for received imaging data based on prescribed behavior. The flexibility that the rules engine provides enables handling every conceivable clinical and research scenario. The destination logical group is the downstream analogue to the source side allowing various connectivity configuration options for downstream systems. Both the source and destination logical groups of functionalities in the router have flexible and extensible options that include normalizing, modifying, and compressing the data. These can all be configured via an intuitive user interface.

The various options supplied by the DICOM Router provide significant power for designing and building AI algorithms with MONAI. In addition to providing the flexibility needed for the design and build phases of AI development, the router may act as an adaptation layer in clinical environments by normalizing the seemingly endless variants of clinical data to an expected dataset for the algorithm. Figure 6 shows this entire pipeline connected.

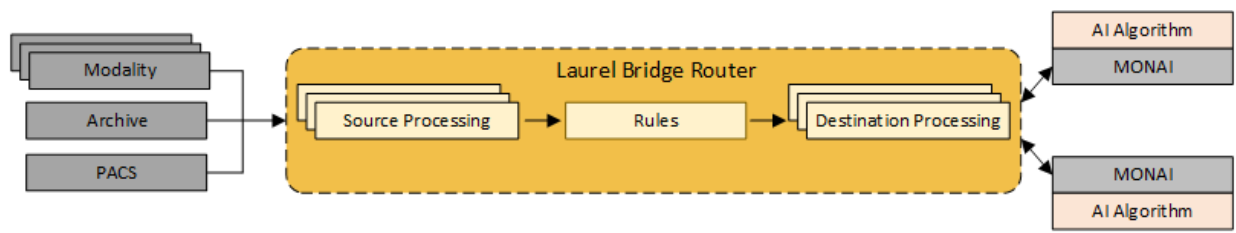

*Figure 6: Connecting MONAI and DICOM Router (Laurel Bridge) together.*

## Conclusion

In this white paper, a successful pathway, utilizing Smart Routing Rules and Workflow Management functionalities of a DICOM router and the MONAI Application Package, has been shown for deploying AI models with medical images within hospital environments. This approach walked through using the AI model inference tool and stepping through the stages involved in the deployment of AI models within hospital settings, all while ensuring compliance with established standards.

The goal for this paper is that readers will be motivated by this implementation and try replicating the workflow in their own environments, leveraging the necessary guidelines, guardrails, and regulations. MONAI, as an open-source community, provides many MONAI Deploy tutorials and examples to the broader community to build their own application package using models shared in the MONAI model zoo or developed in-house. Even though models are trained and vetted by the industry experts and through peer-reviewed research articles, it is important to note that accuracy may vary for a variety of factors. Techniques like Human-In-The-Loop (HITL) processes for adjudicating the AI results should be considered.